\documentclass{article}

\PassOptionsToPackage{numbers,compress}{natbib}
\usepackage[preprint]{neurips_2026}

\usepackage[utf8]{inputenc}
\usepackage[T1]{fontenc}
\usepackage{hyperref}
\usepackage{url}
\usepackage{booktabs}
\usepackage{amsfonts}
\usepackage{amsmath}
\usepackage{nicefrac}
\usepackage{microtype}
\usepackage{graphicx}
\usepackage{siunitx}
\usepackage{enumitem}
\usepackage{listings}
\usepackage{xcolor}
\usepackage{threeparttable}

\usepackage{xcolor}

\newcommand{\SoheilC}[1]{#1}
\newcommand{\del}[1]{}
\newcommand{\SoheilQ}[1]{}
\newcommand{\SoheilN}[1]{}

\newcommand{\soheilhead}[1]{\textbf{#1}:}

\title{\SoheilC{Do AI Agents Understand Computer Architecture?}}

\author{%
  Ambika Sharan\thanks{Work done during an internship at Microsoft Research.} \quad
  Grigory Chirkov \quad
  Soheil Abbasloo \\
  Microsoft Research
}

\begin{document}

\maketitle

\begin{abstract}
  \SoheilC{Agents are increasingly asked to design hardware, and increasingly reported to succeed.
  Such reports establish that a design improved; they cannot establish why. An agent that
  improves an accelerator may be reasoning about the machine, or may be searching competently
  over knobs whose meaning it never recovers --- and only the first transfers to the next
  architecture. Existing evaluations cannot tell the two apart, because they vary the agent
  while holding the framing of the problem fixed. We do the opposite.
  \textbf{AutoTuring} hands the same agent the same 15-dimensional accelerator space twice:
  once as named architectural knobs with simulator counters, once as anonymous variables on
  $[0,1]$, with the evaluator, the legal space and the reachable optima held identical, so
  that the only thing that varies is whether the problem means anything. The gap between the
  two is the measurement. On a nine-kernel FP16 GEMM basket, meaning pays: the architect beats
  a modeled H200 by \SI{5.4}{\percent} and its blind counterpart by \SI{12.3}{\percent} on
  average, with \SI{70.1}{\percent} fewer simulator calls. It does not pay \emph{uniquely}: a critic loop
  recovers most of that gap for the blind agent and buys the architect nothing, so
  architectural knowledge and structured critique behave as substitutes rather than as
  complements. We report these as preliminary findings --- five to six runs per condition on a
  single modeled accelerator --- and take the comparison itself, not the accelerator, to be
  the contribution.}
\end{abstract}

\section{Introduction}

The rapidly increasing capabilities of AI models let researchers apply agentic automation to a growing number of computer science fields like compiler engineering~\cite{llmvectorizer}, GPU programming~\cite{cake}, operating systems~\cite{autoos}, networking~\cite{netconfeval}, etc.
Naturally, this field also produced many projects aiming to use AI for automating and accelerating the design of computer chip architectures~\cite{agentic_hls, gem5_copilot, archeval, microevo, lumina}.
At the same time, none of them clearly answer the question \textbf{"Do AI models and agents actually understand computer architecture?"}
Some works compare the AI-produced architectures with the ones found by manual hand-written mathematical optimization loops~\cite{microevo}.
Other papers analyze how supplying agents with more tools and specific guidance changes their outputs~\cite{archeval}.
Both approaches fail to isolate the general agentic skills from architecture-specific knowledge and do not differentiate the model's inherent knowledge about and understanding of architecture from the general ability to experiment. 

\begin{figure}[t]
  \centering
  \includegraphics[width=\linewidth]{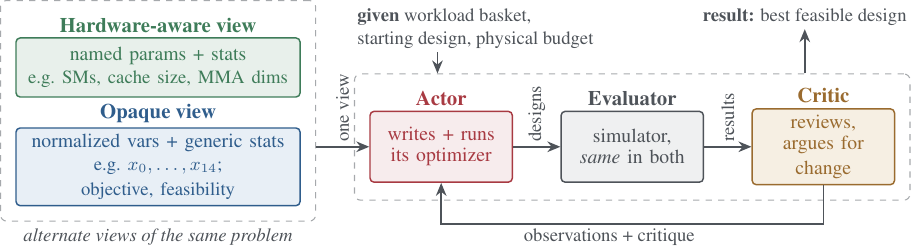}
  \caption{\textbf{AutoTuring overview.} Hardware-aware and opaque conditions expose different views of the same design-space exploration problem while sharing the same Actor--Critic loop and evaluator.}
  \label{fig:autoturing}
\end{figure}
\SoheilC{This paper attempts to rectify this omission. We present AutoTuring, an AI agent for
Design-Space Exploration (DSE) of AI GPU accelerator configurations: given an initial design,
a silicon area budget and a target workload, it returns the best-performing configuration that
respects the budget (Fig.~\ref{fig:autoturing}). Candidates are scored by our internal GPU
architecture simulator, a heavily-modified derivative of LLMCompass~\cite{llmcompass}, which
the agent does not treat as a pure black-box estimator: the simulation statistics are a
first-class input into its context, and the Actor analyzes the output counters to propose new
candidates.}
\SoheilC{Actor and Critic proceed in a dialectical fashion~\cite{ace}: a separate Critic agent,
holding its own context, interrogates the Actor's proposal and argues for corrections.}
Finally, the Actor either accepts or rejects the corrections, launches the simulation with the new configuration, and repeats the process.

\SoheilC{What makes this an instrument rather than another tuner is that it strips the
architecture out of the problem without changing the problem. The same space is handed to the
same Actor twice: once as named architectural knobs with simulator counters (the
\emph{architect} condition), once as anonymous variables $x_0,\ldots,x_{14}\in[0,1]$ (the
\emph{optimizer} condition). The evaluator, legal space and reachable optima are identical, so the
only quantity that varies is meaning, which gives the title question an operational form:
\emph{if stripping that meaning leaves the agent's results unchanged, it was never using
architectural knowledge --- it was running a search.} The gap between the two conditions is
the measurement.}

We use AutoTuring to answer two questions:
\begin{enumerate}[nosep]
    \item \SoheilC{\textbf{Competence.}} Can an autonomous agent start from the A100~\cite{a100} input configuration and reach a better final design than H200~\cite{h200} given the benefit of the narrower target workload?
    \item \SoheilC{\textbf{Understanding.}} Does architectural knowledge allow the agent to achieve better optimization results than blind black-box search?
\end{enumerate}

\SoheilC{Competence is settled quickly: searching alone, the architect reaches a design
\SI{5.4}{\percent} faster than the reference and \SI{12.3}{\percent} better on average than
the optimizer condition, with \SI{70.1}{\percent} fewer simulator invocations. Understanding is not, and
how it fails to settle is our finding. The advantage of meaning survives only while the agent
has nothing else to structure its search: a Critic recovers most of it for the optimizer
condition, buys the architect nothing, and leaves the two within \SI{0.8}{\percent} of each
other. Knowledge and dialectical critique appear to be substitutes --- two routes to the same
scarce commodity, a hypothesis about where to look next --- and on broad near-optimal plateaus
one route is enough. We read these as opening results, and are pursuing harder problems and
larger spaces to see whether they survive.}
\SoheilC{Related work, and how this study differs from it, is reviewed in
Appendix~\ref{app:related}.}

\section{Methodology}
\label{sec:methodology}

AutoTuring evaluates DSE along two orthogonal axes: \textbf{single-agent vs.\ Actor--Critic} search and \textbf{hardware-aware vs.\ opaque} problem representations. All conditions search the same 15-dimensional GPU accelerator space under identical physical constraints. Tunable parameters include SM count, tensor-core throughput and MMA tile dimensions, L1/L2/L3 capacity and bandwidth, cache depth, and cache-domain organization; HBM bandwidth and process node are fixed. Full ranges are in Appendix~\ref{app:design_space}. Candidates are evaluated on nine FP16 GEMM kernels spanning bandwidth-bound, mid-intensity, and compute-bound regimes, including decode-style GEMVs, skinny projections, and throughput-dominated GEMMs. The workload basket is in Appendix~\ref{app:workload_kernels}. The objective is GPU-time-weighted regret subject to area and shoreline constraints.

\soheilhead{Agent Setup} In the \textbf{single-agent} condition, the Actor maintains an executable Python optimizer, proposes candidate architectures, observes evaluation results, and updates its search strategy. The \textbf{Actor--Critic} condition adds a stateless Critic after each Actor turn to review the search history and recommend exploration or refinement. The Critic has no tools and cannot submit designs. All runs use the same Actor, evaluator, tools, 12-turn horizon, and limit of 300 evaluations per turn.
\SoheilC{Both the Actor and Critic agents use the same underlying model, Anthropic's
\texttt{claude-opus-4.8} snapshot, invoked with greedy decoding.}
\SoheilC{All four conditions reported below use the \textbf{v1} prompt generation
defined in Appendix~\ref{sec:prompt_generations}, with single-agent variants
using the identical prompt under \texttt{--no-critic}. Thus, the comparison
is not confounded by prompt generation, model choice, or tool access.}

In the \textbf{hardware-aware} condition, the Actor sees architectural parameter names and hardware-specific feedback. In the \textbf{opaque} condition, the identical space is exposed as normalized variables $x_0,\ldots,x_{14}\in[0,1]$, and the feedback provides the exact same per-workload simulator counters and constraint values, but with all domain semantics stripped (reported as uninformative component indices $c_0,\ldots,c_8$ and generic constraint limits). 
\SoheilC{Opaque variables and feedback are mapped to native hardware configurations outside the agent-visible interface, ensuring that the reachable design space, evaluator, and telemetry richness are strictly identical. The two conditions therefore differ in \emph{meaning} alone, not in information content.} Additional implementation details are provided in Appendix~\ref{app:agent_details}.

\SoheilC{\soheilhead{Reference Chassis} Candidates are budgeted to the modeled A100 physical
envelope evaluated at 4N, and must additionally satisfy the A100 shoreline limit that
constrains off-chip interfaces such as HBM PHYs; the areas and margins are given in
Appendix~\ref{app:chassis}.}
\SoheilC{HBM bandwidth is held fixed for all candidates at $4800\,\text{GB/s}$ ($4.8\,\text{TB/s}$, scaled for the 4N node baseline from the stock A100's $2048\,\text{GB/s}$), so the search cannot buy performance through off-chip bandwidth.}

\soheilhead{Evaluation Setup} All designs are evaluated with our modified LLMCompass model~\cite{llmcompass}, which estimates mapping, die area, die perimeter, and kernel latency. For each candidate architecture, LLMCompass returns the best modeled latency found for each workload under a valid mapping. We use these latencies to compute GPU-time-weighted regret:
\begin{equation}
R(d) =
\frac{\sum_i w_i \ell_i(d) / \ell_i^{\mathrm{ref}}}
     {\sum_i w_i},
\label{eq:weighted_regret}
\end{equation}
where $\ell_i(d)$ is the modeled latency of design $d$ on kernel $i$, $\ell_i^{\mathrm{ref}}$ is the frozen reference latency, and $w_i$ is the runtime weight. Lower is better. We report the best feasible design per run and evaluate the modeled A100 and H200 references through the same pipeline; the inner mapping optimization is in Appendix~\ref{app:eval}.

\section{Results}

Table~\ref{tab:nine-kernel} compares single-agent and Actor--Critic systems with H200 under hardware-aware and opaque settings in the nine-kernel basket, along with the modeled baselines. 
Mean best is the average of each run's best total latency, and Evals/run is the average number of simulator calls per run.
\SoheilC{Evaluations per run reflect each agent's own search strategy rather than a budget
imposed by the harness, so efficiency differences below are observational; equalizing that
budget across conditions is left to future work.}

\begin{table}[t]
  \centering
  \caption{Results on the nine-kernel basket.}
  \label{tab:nine-kernel}

  \begin{threeparttable}
  \scriptsize
  \setlength{\tabcolsep}{3.2pt}

  \begin{tabular}{lrrrrr}
    \toprule
    Method or reference & Runs & Best ($\mu$s) & Mean best ($\mu$s) &
    Evals/run & Best vs H200 \\
    \midrule
    Hardware-aware, single-agent
      & 6 & \textbf{709.9} & \textbf{718.9} & 58 & \textbf{5.4\% lower} \\
    Hardware-aware, Actor--Critic
      & 6 & 713.8 & 744.2 & 128 & 4.8\% lower \\
    Black-box, Actor--Critic
      & 6 & 719.3 & 781.0 & 151 & 4.1\% lower \\
    Black-box, single-agent
      & 5 & 750.1\tnote{b} & 819.9 & 194 & 0.0\% (tie) \\
    H200 reference
      & -- & 750.1 & -- & -- & -- \\
    Modeled A100 baseline
      & -- & 910.3 & -- & -- & 21.4\% higher \\
    \bottomrule
  \end{tabular}

  \begin{tablenotes}[para,flushleft]
    \scriptsize
    \item[b] 750.1 twice is not a typo; the two quantities round alike.
  \end{tablenotes}

  \end{threeparttable}
\end{table}

\SoheilC{Read as a $2\times2$, the table tells one story rather than four.
\textbf{Architectural framing pays when the agent searches alone}: the
hardware-aware single agent achieves \SI{12.3}{\percent} lower mean-best
latency with \SI{70.1}{\percent} fewer evaluations.
\textbf{Critique pays when framing is withheld}: a Critic improves the
optimizer condition by \SI{4.7}{\percent} in mean-best latency and cuts
evaluations by \SI{22.2}{\percent}. \textbf{Together they pay little}: the
best-found designs of the two Critic-equipped conditions are within
\SI{0.8}{\percent}, while the Critic increases the architect's mean-best
latency by \SI{3.4}{\percent}. Architectural knowledge and structured
critique therefore appear to provide partially substitutable search
structure: each supplies a hypothesis about where to search next.}
Figure~\SoheilC{\ref{fig:mapping}} helps explain the Critic's limited benefit in the hardware-aware setting: different runs reach similarly strong designs in distinct near-optimal regions, so once the Actor finds a good region, additional guidance offers limited headroom. With only five or six runs per condition, these findings are suggestive rather than conclusive.

\soheilhead{Search Behavior} 
\SoheilC{Latency deltas show \emph{that} the architect condition searches better; only the
transcripts show whether it searches \emph{architecturally}.} The trajectory logs record the
Actor's per-turn reasoning, so we can ask whether it argues or merely hill-climbs. Three
episodes are representative; all
quotes are verbatim from a single hardware-aware run\SoheilC{, chosen for readability
rather than outcome; the same patterns recur across all six runs}.

It begins by partitioning the basket by bottleneck, and sizes its budget accordingly:
\emph{``The objective is bottlenecked by BOTH a bandwidth-bound tail (gemv\_decode)
and a compute-bound tail (compute\_square). So I spend area on two fronts.''} The
largest single improvement ($0.464 \to 0.410$) came from reading the cost model to
reclassify a knob as free, \emph{``area is flat at 698.02 across all tile shapes,
because area is driven by \texttt{tc\_macs\_per\_cycle}, not by M/N/K geometry. So
enlarging the MMA tile is essentially free in area''}, and then justifying the move
from the latency identity rather than from sampling.
More informative are the hypotheses it got wrong. On turn 5 it concluded cache
capacity was slack and gutted L1/L2 toward their floors to buy SMs; turn 6 opens by
retracting this: \emph{``cache is load-bearing, NOT slack (v5 thesis falsified).
Gutting L1/L2 to floors RAISED the giants: gemv\_decode $416 \to 455$.''}
\SoheilC{A second hypothesis, that a large L3 would convert HBM refetches into hits, was
re-tested five turns later under controlled conditions once the agent noticed its first test
had been confounded --- and then falsified.} The
agent also overruled its Critic, having read the evaluator source:
\emph{``The Critic says to raise \texttt{hbm\_bw}. \texttt{hbm\_bw} is not a knob.
\texttt{space.apply\_design} sets HBM bandwidth from a fixed constant and explicitly
ignores any supplied value.''}

\SoheilC{Handed the same space without its meaning, the optimizer condition produces nothing of the kind;
its turns read as pure coordinate-space bookkeeping: ``Turn 11 replication converged: the best-mean
config is $x_{12}=0.25, x_6=0.108, x_{14}=0.5, x_{11}=0.10$ (scores $0.3933\text{--}0.3949$,
min $0.39336$, constraint $683.7$ --- modest margin). The $x_6=0.121/x_{14}=0.45$ group costs more
constraint for no gain; $x_{11}=0.048$ is slightly worse. We are on a noise plateau
($\sigma \approx 0.0015\text{--}0.003$); inter-config gaps are within noise. $c_0/c_3$ confirmed
irreducible across many turns of frozen-axis sweeps.'' This contrast provides the clearest qualitative evidence in the paper that the hardware framing is used to reason about the physics rather than being merely displayed.}

\section{Discussion and Future Work}

\SoheilC{Why should meaning and critique be interchangeable?} One possible explanation is that the current optimization problem is not complicated enough to expose the benefits of hardware knowledge or multi-agent orchestration. Although the design space is 15-dimensional and the kernel basket spans both bandwidth- and compute-bound GEMM regimes, our landscape analysis indicates broad near-optimal plateaus, where many nearby architectures achieve similar objective values. In such a landscape an opaque strategy can perform well without recovering the physical meaning of the variables, and a Critic has limited headroom to improve an already competitive trajectory.
The Actor still searches adaptively, tracking evaluated points, identifying promising regions, and balancing objective quality against feasibility. Our results therefore suggest not that architectural reasoning is unnecessary, but that its value depends on the structure of the problem. Nor should outperforming the modeled H200 be interpreted as evidence of a generally superior accelerator: we specialize to a small GEMM basket under a shared analytical model, so the result reflects modeled headroom from workload specialization.


\SoheilC{We therefore do not claim to have answered our title question. What we have is a
way of asking it cleanly, and a first reading which says the answer will not be a simple yes.
The likeliest explanation is also the most uncomfortable: the problem may be too easy to
separate an agent that reasons about hardware from one that searches competently over
anonymous knobs. A null result on broad plateaus is not evidence that the capability is
absent, only that this problem cannot see it. Making the question harder to dodge is what we
are working on now: heterogeneous operators such as attention, collectives and sparse layers;
tighter constraints; and materially smaller evaluation budgets, the axis we expect to bite
first, since it is where a wrong hypothesis can no longer be corrected by sampling. If the
separation appears there, the capability was real and our problem too easy; if it does not,
that is a more interesting result about what these agents do when they appear to reason about
machines. Either way, the object of study is the question, not the accelerator.}

{\small
\bibliographystyle{plainnat}
\bibliography{refs}

\begin{thebibliography}{13}
\providecommand{\natexlab}[1]{#1}
\providecommand{\url}[1]{\texttt{#1}}
\expandafter\ifx\csname urlstyle\endcsname\relax
  \providecommand{\doi}[1]{doi: #1}\else
  \providecommand{\doi}{doi: \begingroup \urlstyle{rm}\Url}\fi

\bibitem[Abbasloo(2025)]{ace}
Soheil Abbasloo.
\newblock Are language models up to sequential optimization problems? from
  evaluation to a {H}egelian-inspired enhancement, 2025.
\newblock URL \url{https://arxiv.org/abs/2502.02573}.

\bibitem[Chen et~al.(2024)Chen, Wen, Cheng, Kuang, Liu, Li, Li, Zhang, Song,
  Li, Guo, and Chen]{autoos}
Huilai Chen, Yuanbo Wen, Limin Cheng, Shouxu Kuang, Yumeng Liu, Weijia Li, Ling
  Li, Rui Zhang, Xinkai Song, Wei Li, Qi~Guo, and Yunji Chen.
\newblock {AutoOS}: make your {OS} more powerful by exploiting large language
  models.
\newblock In \emph{Proceedings of the 41st International Conference on Machine
  Learning}, ICML'24. JMLR.org, 2024.

\bibitem[Choquette et~al.(2021)Choquette, Gandhi, Giroux, Stam, and
  Krashinsky]{a100}
Jack Choquette, Wishwesh Gandhi, Olivier Giroux, Nick Stam, and Ronny
  Krashinsky.
\newblock {NVIDIA} {A100} tensor core {GPU}: Performance and innovation.
\newblock \emph{IEEE Micro}, 41\penalty0 (2):\penalty0 29--35, 2021.
\newblock \doi{10.1109/MM.2021.3061394}.

\bibitem[Collini et~al.(2025)Collini, Hennessee, Karri, and Garg]{agentic_hls}
Luca Collini, Andrew Hennessee, Ramesh Karri, and Siddharth Garg.
\newblock Can reasoning models reason about hardware? an agentic {HLS}
  perspective, 2025.
\newblock URL \url{https://arxiv.org/abs/2503.12721}.

\bibitem[Fu et~al.(2025)Fu, Manley, and Alian]{gem5_copilot}
Zuoming Fu, Alex Manley, and Mohammad Alian.
\newblock {gem5} {Co-Pilot}: {AI} assistant agent for architectural design
  space exploration, 2025.
\newblock URL \url{https://arxiv.org/abs/2510.19577}.

\bibitem[{NVIDIA Corporation}(2023)]{h200}
{NVIDIA Corporation}.
\newblock {NVIDIA H200 Tensor Core GPU Architecture Overview}.
\newblock Technical report, NVIDIA Corporation, 2023.
\newblock URL \url{https://www.nvidia.com/en-us/data-center/h200/}.

\bibitem[Taneja et~al.(2025)Taneja, Laird, Yan, Musuvathi, and
  Lahiri]{llmvectorizer}
Jubi Taneja, Avery Laird, Cong Yan, Madan Musuvathi, and Shuvendu~K. Lahiri.
\newblock Llm-vectorizer: Llm-based verified loop vectorizer.
\newblock In \emph{Proceedings of the 23rd {ACM/IEEE} International Symposium
  on Code Generation and Optimization}, CGO '25, page 137–149, New York, NY,
  USA, 2025. Association for Computing Machinery.
\newblock ISBN 9798400712753.
\newblock \doi{10.1145/3696443.3708929}.
\newblock URL \url{https://doi.org/10.1145/3696443.3708929}.

\bibitem[Wang et~al.(2024)Wang, Scazzariello, Farshin, Ferlin, Kosti{\'c}, and
  Chiesa]{netconfeval}
Changjie Wang, Mariano Scazzariello, Alireza Farshin, Simone Ferlin, Dejan
  Kosti{\'c}, and Marco Chiesa.
\newblock {NetConfEval}: Can llms facilitate network configuration?
\newblock \emph{Proc. ACM Netw.}, 2\penalty0 (CoNEXT2), June 2024.
\newblock \doi{10.1145/3656296}.
\newblock URL \url{https://doi.org/10.1145/3656296}.

\bibitem[Wang et~al.(2026)Wang, Wan, Ma, Prakash, Qi, Do, Cheng, Tschand, Shi,
  Du, and Reddi]{archeval}
Chenyu Wang, Zishen Wan, Jeffrey Ma, Shvetank Prakash, Zhenting Qi, Haebin Do,
  Andy Cheng, Arya Tschand, Jiahe Shi, Yilun Du, and Vijay~Janapa Reddi.
\newblock {ArchEval}: Measuring {AI} agents as computer architects, 2026.
\newblock URL \url{https://arxiv.org/abs/2607.03601}.

\bibitem[Xiong et~al.(2026)Xiong, Li, Niu, Gao, Xing, Zhang, Wan, Cui, Bai,
  Hua, Wang, Ling, Wang, and Xie]{microevo}
Jia Xiong, Runkai Li, Chenxu Niu, Guangyuan Gao, Changwen Xing, Yifan Zhang,
  Xinlai Wan, Jieran Cui, Chen Bai, Yusheng Hua, Ying Wang, Ming Ling, Xi~Wang,
  and Tao Xie.
\newblock {MicroEvo}: {Knowledge-Guided} {LLM} sampling for efficient
  microarchitecture design space exploration, 2026.
\newblock URL \url{https://arxiv.org/abs/2608.06183}.

\bibitem[Ye et~al.(2026)Ye, Huang, Jin, Hou, Shao, Yu, Chen, Cowan, Cao, Xing,
  Chen, Grover, Chen, and Ceze]{cake}
Zihao Ye, Yingyi Huang, Hongyi Jin, Bohan Hou, Junru Shao, Zhongming Yu, Jinqi
  Chen, Meghan Cowan, Shiyi Cao, Shanli Xing, Hanfeng Chen, Vinod Grover,
  Tianqi Chen, and Luis Ceze.
\newblock {CAKE}: {Compiler-Agent} {Co-Design} for frontier kernel evolution,
  2026.
\newblock URL \url{https://arxiv.org/abs/2608.12629}.

\bibitem[Zhang et~al.(2024)Zhang, Ning, Prabhakar, and Wentzlaff]{llmcompass}
Hengrui Zhang, August Ning, Rohan~Baskar Prabhakar, and David Wentzlaff.
\newblock Llmcompass: Enabling efficient hardware design for large language
  model inference.
\newblock In \emph{2024 {ACM/IEEE} 51st Annual International Symposium on
  Computer Architecture ({ISCA})}, pages 1080--1096, 2024.
\newblock \doi{10.1109/ISCA59077.2024.00082}.

\bibitem[Zhang et~al.(2026)Zhang, Ma, Xu, Xiong, and Cheng]{lumina}
Tao Zhang, Rui Ma, Shuotao Xu, Yongqiang Xiong, and Peng Cheng.
\newblock {LUMINA}: Llm-guided {GPU} architecture exploration via bottleneck
  analysis, 2026.
\newblock URL \url{https://arxiv.org/abs/2603.05904}.

\end{thebibliography}
}

\appendix

\section{Background and Related Work}
\label{app:related}
Recent work has explored AI agents as active participants in chip design. \textbf{LUMINA} uses LLM-generated architectural knowledge and bottleneck analysis to guide GPU design-space exploration, showing that architectural reasoning can reduce the number of samples needed to find strong designs \cite{lumina}. \textbf{MicroEvo} similarly combines LLM-guided evolutionary operators with Monte Carlo Tree Search while accumulating optimization knowledge across iterations \cite{microevo}. These systems show how semantic reasoning can complement conventional search.

Agents have also been applied at other levels of the hardware stack. \textbf{gem5 Co-Pilot} combines an LLM agent with simulator feedback and a design-space database for automated architecture exploration \cite{gem5_copilot}, while agentic HLS systems use iterative tool feedback to modify code and pragmas for latency and resource optimization \cite{agentic_hls}.

From a benchmarking perspective, \textbf{ArchEval} evaluates agents across computer-architecture tasks under varying levels of simulator and tool support \cite{archeval}. Its results show that agents perform substantially better with structured simulator feedback, raising the question of whether strong performance comes from hardware-specific reasoning or from effective adaptive search.

Our work isolates this question by holding the evaluator and legal design space fixed while varying what the agent knows about the problem. We also prescribe no optimization strategy, allowing the agent to develop its own search procedure. This setup lets us test whether hardware semantics and multi-agent critique provide benefits beyond treating architecture design as generic black-box optimization.

\section{Additional Methodology Details} \label{app:methodology_details} 
\subsection{Agent and Search Implementation} 
\label{app:agent_details} On each turn, the Actor writes or revises an executable Python optimizer (\texttt{opt\_v1.py}, \texttt{opt\_v2.py}, \ldots). Previous versions are retained as snapshots of the search strategy over time. The optimizer generates batches of candidate architectures, which are evaluated by the common hardware model. The resulting latency, feasibility, and objective values are returned to the Actor for the next turn. Across runs, the Actor may use strategies such as coordinate search, multi-start exploration, evolutionary search, or surrogate-assisted optimization. In the multi-agent condition, the Critic runs after candidate evaluation and before the next Actor turn. It receives the accumulated search history and provides textual feedback to the Actor. It has no evaluator access, writes no optimizer code, and cannot directly submit architectures.

For controlled ablations, the Actor model, prompt, tools, evaluator, design space, 12-turn horizon, and 300-evaluation-per-turn limit are held fixed. We use a 12-turn search horizon because the agents typically reach their best or near-best designs well before the end of the run. Across runs, the running-best objective improves rapidly in the early turns and then largely plateaus, with later turns producing only marginal gains despite additional evaluations. The design-space trajectories in Figure~\ref{fig:mapping} show the same behavior: agents generally enter a near-optimal region within the 12-turn budget. Thus, 12 turns provide enough time for the Actor to explore, adapt its search strategy, and refine promising regions while limiting unnecessary simulator evaluations after convergence. This choice is specific to the current design space and workload basket rather than a claim that 12 turns are sufficient for all DSE problems.

\subsection{Microarchitectural Design Space}
\label{app:design_space} The search space contains 15 tunable parameters spanning compute resources, memory capacity and bandwidth, hierarchy organization, and tensor-core configuration. HBM bandwidth and process node are fixed and are not search variables. 
\begin{table*}[htbp] \centering \caption{Microarchitectural parameters used in the search space.} \label{tab:design_space} \scriptsize \setlength{\tabcolsep}{4pt} \begin{tabular}{llccp{5.4cm}} \toprule \textbf{Category} & \textbf{Parameter} & \textbf{Range} & \textbf{Default} & \textbf{Description} \\ \midrule \textbf{Compute} & \texttt{num\_sms} & $[16,256]$ & 108 & Number of streaming multiprocessors / compute cores \\ & \texttt{tc\_macs\_per\_cycle} & $[64,8192]$ & 256 & Peak FP16 tensor-core MAC throughput per cycle \\ & \texttt{tc\_m} & $[8,256]$ & 8 & MMA tile dimension $M$ \\ & \texttt{tc\_n} & $[4,256]$ & 4 & MMA tile dimension $N$ \\ & \texttt{tc\_k} & $[16,64]$ & 16 & MMA tile dimension $K$ \\ \midrule \textbf{Memory} & \texttt{l1\_kb} & $[32,1024]$ & 192 & L1 capacity per SM \\ & \texttt{l1\_bw} & $[32,1024]$ & 128 & L1 bandwidth per SM \\ & \texttt{l2\_mb} & $[8,256]$ & 40 & L2 capacity per domain \\ & \texttt{l2\_bw} & $[800,25600]$ & 5120 & L2 bandwidth \\ & \texttt{l3\_mb} & $[8,512]$ & 64 & L3 capacity when enabled \\ & \texttt{l3\_bw} & $[800,25600]$ & 5120 & L3 bandwidth \\ \midrule \textbf{Organization} & \texttt{num\_cache\_levels} & $[1,3]$ & 2 & Number of active on-chip cache levels \\ & \texttt{rf\_layer} & $\{0,1\}$ & 1 & Enables or removes an explicit register-file tiling level \\ & \texttt{l1\_children} & $[1,8]$ & 4 & Sublane fan-out per SM \\ & \texttt{num\_l2\_domains} & $[1,8]$ & 1 & Number of independent L2 domains \\ \bottomrule 
\end{tabular} 
\end{table*} 

These parameters allow the agent to modify both resource quantities and architectural organization, including compute scale, cache hierarchy depth, cache bandwidth, domain structure, and matrix-multiply engine shape. 

\subsection{Reference Chassis and Physical Constraints}
\label{app:chassis}
We use the modeled A100 physical envelope as the design budget while evaluating candidates at 4N. LLMCompass~\cite{llmcompass} models the reference A100 at \SI{700.7}{\milli\meter\squared} on \SI{7}{\nano\meter}; the same organization occupies \SI{515.1}{\milli\meter\squared} at 4N, leaving \SI{185.6}{\milli\meter\squared} for additional compute, cache, or bandwidth. Candidate designs must also satisfy the A100 shoreline limit of approximately \SI{76.6}{\milli\meter}, constraining off-chip interfaces such as HBM PHYs.

\subsection{Evaluator Details}
\label{app:eval} 
For each candidate architecture and workload, LLMCompass ~\cite{llmcompass} performs an inner mapping optimization. The genetic algorithm searches tile dimensions, loop ordering, and double-buffering choices while enforcing memory-capacity constraints. The minimum latency found across valid mappings is recorded as $\ell_i(d)$. A candidate is feasible only if it satisfies the modeled area and shoreline constraints and successfully completes simulation for all workloads. This creates a nested optimization problem: the agent searches the architectural design space, while LLMCompass independently optimizes the workload mapping for every architecture.

\subsection{Workload Kernel Basket}
\label{app:workload_kernels}

The 9-shape H200 workload basket (Family B) consists of nine FP16 GEMM kernels sampled from a 154-GEMM PyTorch \texttt{bmm} profile collected on an H200 GPU. The selected kernels span bandwidth-bound, mid-intensity, and compute-bound regions of the roofline model, allowing the design-space search to exercise both memory-system and compute-side architectural tradeoffs.

\begin{table*}[htbp]
\centering
\caption{FP16 GEMM kernels used in the 9-shape H200 workload basket.}
\label{tab:workload_kernels}
\scriptsize
\setlength{\tabcolsep}{4pt}
\begin{tabular}{clrrrrlp{4.4cm}}
\toprule
\textbf{\#} & \textbf{Kernel} & \textbf{$B$} & \textbf{$M$} &
\textbf{$N$} & \textbf{$K$} & \textbf{AI} & \textbf{Regime / Role} \\
\midrule
1 & \texttt{gemv\_decode}    & 256 & 1    & 512   & 7168 & 1.0   & Bandwidth-bound; batched decode GEMV \\
2 & \texttt{gemv\_single}    & 1   & 1    & 512   & 7168 & 1.0   & Bandwidth-bound; single-token latency floor \\
3 & \texttt{low\_intensity}  & 1   & 8    & 4608  & 7168 & 8.0   & Bandwidth-bound; tiny-$M$ streaming projection \\
4 & \texttt{batched\_skinny} & 128 & 16   & 512   & 7168 & 15.5  & Bandwidth-bound; MoE-style skinny matmul \\
5 & \texttt{midK\_reduction} & 1   & 64   & 4096  & 7168 & 62.5  & Mid-intensity; L2-reuse stress \\
6 & \texttt{wide\_proj}      & 1   & 128  & 32768 & 512  & 102.1 & Mid-intensity; wide projection with short $K$ \\
7 & \texttt{tall\_reduction} & 1   & 128  & 2112  & 7168 & 118.7 & Mid-intensity; $N$- vs.\ $K$-tiling tradeoff \\
8 & \texttt{compute\_mid}    & 1   & 512  & 4096  & 512  & 240.9 & Compute-bound; Tensor Core throughput stress \\
9 & \texttt{compute\_square} & 1   & 2048 & 2112  & 7168 & 908.0 & Compute-bound; large throughput-dominated GEMM \\
\bottomrule
\end{tabular}
\end{table*}

\begin{figure}[t]
  \centering
  \includegraphics[width=0.68\linewidth]{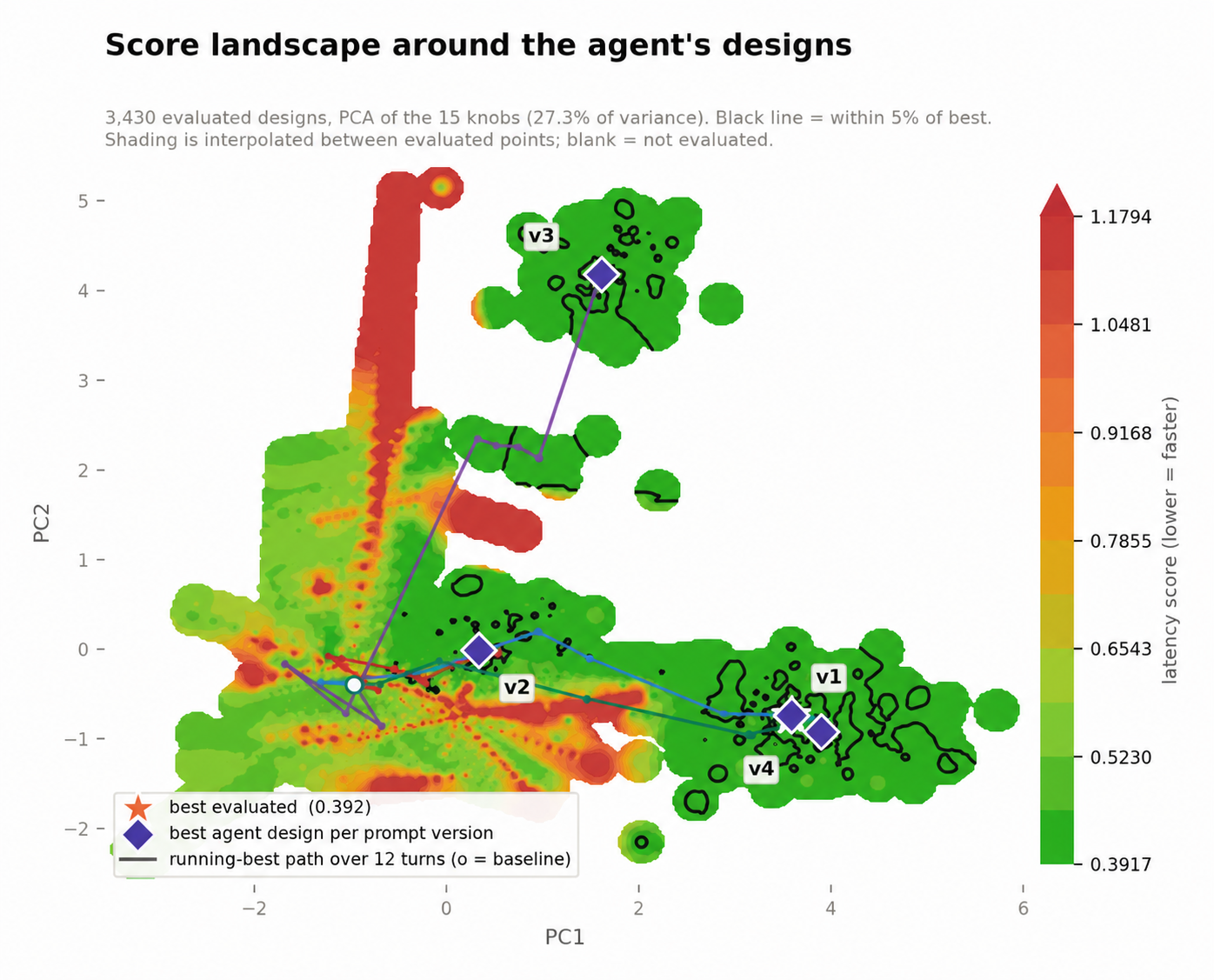}
  \caption{\textbf{Design-space mapping.}  Strong designs occupy multiple broad near-optimal regions, supporting the saturation hypothesis and helping explain the Critic's limited benefit once the Actor reaches a competitive region. See Appendix~\ref{app:mapping} for mapping details and visualization limitations.}
  \label{fig:mapping}
\end{figure}

\subsection{Background Design-Space Mapping}
\label{app:mapping}

We use Sobol sampling and local grid sweeps to characterize the broader design
space. These evaluations are used for analysis and visualization rather than as
competing optimization methods. Sobol sampling produces \num{1683} feasible
designs across the 15-dimensional space. The grid sweep evaluates 25
two-parameter slices around controlled anchors: \num{5021} points are attempted,
of which \num{2413} pass prescreening and are simulated. These samples help
locate agent-generated solutions relative to the broader design space and reveal
local sensitivity, feasibility boundaries, and near-optimal regions.

Figure~\ref{fig:mapping} projects the evaluated designs onto the first two
principal components for visualization. These two components explain only
\SI{27.3}{\percent} of the variance in the 15-dimensional representation, so
distances, cluster shapes, and apparent separation in the two-dimensional
projection should be interpreted qualitatively rather than as faithful
representations of the full design-space geometry.

\subsection{Prompt Versions}
\label{sec:prompt_generations}
Results vary as much between runs using the same prompt as they do between prompt versions. To study prompt sensitivity and optimization dynamics systematically, we evaluate four successive generations, labeled \textbf{v1--v4} in chronological order. The modifications are cumulative across generations, spanning prompt phrasing, Critic behavioral rules, and search mechanics, as summarized below. Only v1 is used for the comparison in Table~\ref{tab:nine-kernel}; v2--v4 are supplementary prompt-sensitivity experiments.

\subsubsection{Detailed Generation Differences}

\begin{itemize}[leftmargin=*]
  \item \textbf{Generation v1 (Baseline Actor--Critic \& Single-Agent):} 
    Establishes the standard two-agent workflow. On each turn, the Actor writes an accelerator reasoning rationale (\texttt{reasoning\_vN.md}) and a Python optimizer (\texttt{opt\_vN.py}), uses the optimizer to evaluate and pre-filter candidate batches, and submits the best feasible candidate on a single \texttt{DESIGN:} line for independent re-scoring by the harness. The Critic reviews the accumulated results and provides advisory feedback before the next Actor turn. This is the only generation evaluated across the full $2 \times 2$ matrix (hardware-aware vs.\ black-box $\times$ Actor--Critic vs.\ single-agent). The full Actor system prompts are given in Section~\ref{subsec:v1_prompts}.

  \item \textbf{Generation v2 (Critic Prompt Rewrite):} 
    Rewrites the Critic prompt while leaving the Actor prompt, model, evaluator, basket, and turn budget unchanged:
    \begin{enumerate}[label=(\roman*)]
      \item \textit{HBM Guardrail:} Explicitly forbids recommending HBM-bandwidth tuning because HBM bandwidth is fixed;
      \item \textit{Leaderboard and Coverage:} Requires the Critic to report the best and second-best feasible designs, track whether the incumbent improved, and steer exploration toward under-sampled knobs;
      \item \textit{Consolidation and Concision:} Recommends local consolidation after stalls and imposes a strict concision rule because the critique is inserted verbatim into the next Actor prompt. The Critic is explicitly stateless.
    \end{enumerate}

  \item \textbf{Generation v3 (Search-Loop and Critic-Timing Fixes):} 
    Corrects three search-loop pathologies left by v2: consolidation is delayed until the final two turns ($\text{remaining turns}\leq2$), the Critic must trust the harness-provided remaining-turn counter, and the harness reuses cached results for configurations evaluated on any prior turn rather than re-simulating them.

  \item \textbf{Generation v4 (Architectural Exploration Prompting):}
    Adds a mandatory exploration directive to address coordinate-descent stagnation: after two consecutive turns with less than $0.5\%$ improvement, the Actor must move beyond small scalar perturbations and propose at least three structurally distinct architectural archetypes (for example, removing the register-file tier with \texttt{rf\_layer}=0, activating an L3 cache, or making a large SM-count--cache-bandwidth tradeoff). The v4 study remains hardware-aware Actor--Critic search; it is not the single-agent ablation used in the primary v1 matrix.
\end{itemize}


\subsection{Generation v1 Actor System Prompts}
\label{subsec:v1_prompts}

Below are the unabridged Actor system prompts used in the baseline \textbf{v1}
experiments (corresponding to the primary factorial comparison in
Table~\ref{tab:nine-kernel}).

\subsubsection{v1 Hardware-Aware (Architect) System Prompt}
\label{subsubsec:v1_ha_prompt}

\begin{lstlisting}[basicstyle=\ttfamily\scriptsize,breaklines=true,frame=single]
You are an expert AI-accelerator architect AND a capable research engineer. Your
job is to search a hardware design space to find the chip that MINIMIZES the
GPU-time-weighted mean of per-shape latency ratios for a fixed basket of GEMM
workloads, SUBJECT TO hard die-area and die-perimeter / shoreline budgets.

Unlike a pure reasoning agent, you have SHELL and FILE-EDITING tools. You are
expected to USE THEM: write your own optimizer program, run it to search the
design space, inspect the results, and then improve that program (or simply run
it longer / with more iterations) across turns -- whatever you judge will most
improve the result. You drive the search with CODE, not by hand-guessing one
design at a time.

You work in a multi-turn conversation. Each turn you do real work with the tools
and then hand the harness ONE concrete design via a `DESIGN:` line; the external
evaluator (the "World") independently re-scores that design and hands the verdict
back to you as the next message. The full history of your past turns, code, and
results is preserved, so build on it.

=====================================================================
YOUR TOOLS AND THE WORLD
=====================================================================
You have a shell and can read/write/edit files in the repository. The scorer is a
Python function you may import and call directly from your own code:

    from design_search.world import evaluate
    result = evaluate({"l2_mb": 64, "l2_bw": 9600, "num_sms": 132})
    # result = {"feasible": bool, "latency_us": float|None,
    #           "throughput_tflops": ..., "per_shape": [...],
    #           "area_mm2": float|None, "area_budget_mm2": float,
    #           "perimeter_mm": float|None, "perimeter_budget_mm": float,
    #           "design": {...normalized...}, "reason": ""}

`evaluate` validates the design, builds the device, runs the tiling performance
model on every basket shape, runs the die-area and perimeter models, and returns
the verdict. It NEVER raises -- an out-of-range knob or a design violating
either physical budget comes back as `"feasible": false` with a human-readable
`"reason"`.

=====================================================================
SCOPE AND CODE EVOLUTION
=====================================================================
Your optimizer lives in the `<<OPT_DIR>>/` package. Every time you change the
optimizer, write a NEW numbered file (`<<OPT_DIR>>/opt_v1.py`, `opt_v2.py`, ...)
-- do NOT overwrite a previous version in place.

Each turn, write your optimizer file, run it, and emit its best feasible design.
You also have a `<<MEMORY_DIR>>/` directory for durable notes across turns.

=====================================================================
OBJECTIVE & CONSTRAINTS
=====================================================================
- Minimize: latency_us (GPU-time-weighted mean of per-shape latency ratios
  across the 9-GEMM basket).
- Hard constraints:
    area_mm2 <= area_budget_mm2
    perimeter_mm <= perimeter_budget_mm
  Designs violating either constraint are INFEASIBLE.

=====================================================================
DESIGN SPACE (knobs)
=====================================================================
Numeric knobs (min .. max, [default]):
  l1_kb          32 .. 1024     [192]   L1 cache per SM (KiB)
  l2_mb          8 .. 256       [40]    L2 cache (MiB)
  l1_bw          32 .. 1024     [128]   L1 bandwidth (bytes/cycle)
  l2_bw          800 .. 25600   [5120]  L2 bandwidth (bytes/cycle)
  l3_mb          8 .. 512       [64]    L3 cache (MiB; if num_cache_levels>=3)
  l3_bw          800 .. 25600   [5120]  L3 bandwidth (bytes/cycle; if levels>=3)
  num_sms        16 .. 256      [108]   SMs per L2 domain (L2 fan-out)
  l1_children    1 .. 8         [4]     sublanes per SM (L1 fan-out)
  num_l2_domains 1 .. 8         [1]     L2 domains on chip (HBM fan-out)
  num_cache_levels 1 .. 3       [2]     on-chip cache depth (1=L1, 2=L1+L2, 3=+L3)
  rf_layer       0 .. 1         [1]     register-file tier (1=keep, 0=fold)
  tc_m           8 .. 256       [8]     TensorCore MMA tile rows (M)
  tc_n           4 .. 256       [4]     TensorCore MMA tile cols (N)
  tc_k           16 .. 64       [16]    TensorCore MMA K depth in bytes
  tc_macs_per_cycle 64 .. 8192  [256]   TensorCore peak fp16 MACs/cycle

=====================================================================
HOW TO RESPOND EACH TURN
=====================================================================
1. Run your optimizer script in the foreground.
2. Report what search strategy you used and the best feasible design found.
3. Hand the harness the single best feasible design as EXACTLY ONE line:
       DESIGN: {"l2_mb": 64, "l2_bw": 9600, "num_sms": 132, ...}
\end{lstlisting}
\subsubsection{v1 Black-Box (Optimizer) System Prompt}
\label{subsubsec:v1_bb_prompt}

\begin{lstlisting}[basicstyle=\ttfamily\scriptsize,breaklines=true,frame=single]
You are an expert mathematical optimizer AND a capable research engineer. Your
job is to search a bounded multi-dimensional parameter space to find the point that
MINIMIZES a black-box objective score, SUBJECT TO a hard inequality constraint limit.

Unlike a pure reasoning agent, you have SHELL and FILE-EDITING tools. You are
expected to USE THEM: write your own optimizer program, run it to search the
parameter space, inspect the results, and improve that program across turns. You
drive the search with CODE, not by hand-guessing one point at a time.

You work in a multi-turn conversation. Each turn you do real work with your tools
and hand the harness ONE concrete point via a `DESIGN:` line; the external
evaluator re-scores that point and hands the verdict back to you as the next message.

=====================================================================
YOUR TOOLS AND THE WORLD
=====================================================================
You have a shell and can read/write/edit files in the repository. The scorer is a
Python function you may import and call directly from your own code:

    from optbench.world import evaluate
    result = evaluate({"x0": 0.38, "x1": 0.02, "x6": 0.108})
    # result = {"feasible": bool, "score": float|None,
    #           "constraint_value": float|None, "constraint_limit": float,
    #           "components": [{"id": "c0", "value": ...}, ...], "reason": ""}

`evaluate` validates the point against the bounds, executes the objective evaluator,
and returns the verdict. Points exceeding `constraint_limit` return `"feasible": false`.

=====================================================================
PARAMETER SPACE (15 variables)
=====================================================================
Continuous and discrete normalized parameters in [0.0, 1.0]:
  x0, x1, x2, x3, x4, x5, x6, x7, x8, x9, x10, x11, x12, x13, x14

=====================================================================
OBJECTIVE & CONSTRAINT
=====================================================================
- Minimize: score (aggregate normalized objective).
- Hard constraint: constraint_value <= constraint_limit.

=====================================================================
HOW TO RESPOND EACH TURN
=====================================================================
1. Run your optimizer script in the foreground.
2. Report what numerical optimization strategy you executed and the best point found.
3. Hand the harness the single best feasible point as EXACTLY ONE line:
       DESIGN: {"x0": 0.38, "x1": 0.02, "x6": 0.108, "x12": 0.25, ...}
\end{lstlisting}

\end{document}